\documentclass{article}
\usepackage{ijcai24}

\usepackage{times}
\usepackage{soul}
\usepackage{url}
\usepackage[hidelinks]{hyperref}
\usepackage[utf8]{inputenc}
\usepackage[small]{caption}
\usepackage{graphicx}
\usepackage{amsmath}
\usepackage{amsthm,amssymb,amsfonts}
\usepackage{booktabs}
\usepackage{algorithm}
\usepackage[switch]{lineno}

\usepackage{natbib}
\usepackage{algpseudocode}
\linenumbers

\title{Compositional Program Generation for Few-Shot Systematic Generalization}
\author{}
\author{
Tim Klinger$^{1,*}$\And
Luke Liu$^{2,*}$\And
Soham Dan,$^1$\And
Maxwell Crouse$^1$\And
Parikshit Ram$^1$\And
Alex Gray$^{1,3}$\\
\affiliations
$^1$IBM Research AI\\
$^2$New York University\\
$^3$Centaur AI Institute\\
\emails
\{tklinger, Soham.Dan, Maxwell.Crouse,  parikshit.ram, alexander.gray\}@ibm.com,
ql2078@nyu.edu
}

\begin{document}
\nolinenumbers
\maketitle

\begin{abstract}
Compositional generalization is a key ability of humans that enables us to learn new concepts from only a handful examples. Neural machine learning models, including the now ubiquitous Transformers, struggle to generalize in this way, and typically require thousands of examples of a concept during training in order to generalize meaningfully. This difference in ability between humans and artificial neural architectures, motivates this study on a neuro-symbolic architecture called the Compositional Program Generator (CPG). CPG has three key features: \textit{modularity}, \textit{composition}, and \textit{abstraction}, in the form of grammar rules, that enable it to generalize both systematically to new concepts in a few-shot manner, as well as productively by length on various sequence-to-sequence language tasks. For each input, CPG uses a grammar of the input language and a parser to generate a parse in which each grammar rule is assigned its own unique semantic module, a probabilistic copy or substitution program. Instances with the same parse are always processed with the same composed modules, while those with different parses may be processed with different modules. CPG learns parameters for the modules and is able to learn the semantics for new rules and types incrementally, without forgetting or retraining on rules it's already seen. It achieves perfect generalization on both the SCAN and COGS benchmarks using just 14 examples for SCAN and 22 examples for COGS -- state-of-the-art accuracy with a 1000x improvement in sample efficiency.
\end{abstract}

\section{Introduction}
One of the long-standing issues with general-purpose neural architectures like Transformers \citep{NIPS2017_3f5ee243}, is that they struggle to learn systematic behavior \cite{ontanon2022making}. For example, a model that has been trained to follow instructions like ``\textit{walk} left twice", ``\textit{run} left twice", and ``\textit{turn} left twice" may fail on the similar ``\textit{jump} left twice" --- even though the system can correctly follow the instruction \textit{jump} by itself. This is often characterized as a failure of \textit{systematicity} since the model seems unable to systematically compose its knowledge of how to perform individual actions like ``jump" and its knowledge of how to ``\underline{\hspace{.5cm}} left twice". Similarly, a model may fail to \textit{productively} generalize to a recursive solution, working for some length inputs but not others. Moreover, one might expect that a model which is able to generalize systematically should be able to learn with less data (few-shot). If it is able to combine its knowledge of things it already knows to solve a new problem during training, then it shouldn't need training data to learn how to solve that problem. Neural networks like Transformers have so far not been able to exploit any systematicity they have to learn with less data.

There have been many different approaches to this problem, some of which we discuss further in the related work. They vary in the kind and amount of additional task-specific information they require (if any). Here we describe one such approach, a neuro-symbolic architecture, Compositional Program Generator (CPG) which compositionally generalizes for sequence-to-sequence language tasks like translation and semantic parsing. It requires a context-free grammar of the input language only and, for the COGS experiments, a dictionary mapping each input word to its interpretation in the output language. On the SCAN benchmark it matches the SOTA but uses less data (14 examples); on the COGS benchmark it achieves a new SOTA, solving it completely using just 22 examples.

CPG has three key features which together support few-shot compositional generalization and incremental learning: 
\begin{enumerate}
    \item \textit{modularity} -- Instead of a monolithic network we use multiple modules with private parameters which can specialize to learn different semantic functions. 
    \item \textit{composition} -- Modules are composed in different configurations for different classes of inputs. This enables combinatorially many different programs to be constructed from a small set of modules.
    \item \textit{abstraction} -- Modules are associated with context-free grammar rules which operate at the level of abstract types, corresponding to classes of input expressions. Each input with the same parse will combine the modules in the same way. This enables whole classes of inputs to be systematically handled by the same composed program, but allows different classes to be handled by different programs.
\end{enumerate}

In the Approach section we discuss in more detail how these features combine to support the learning of a systematic, productive model for sequential language tasks and provide examples from the SCAN and COGS benchmarks. More details of the implementation and a curricular training algorithm are provided in the Model and Curricular Training sections. Experiments with the standard SCAN and COGS benchmarks are described in the Experiments section. Related Work discusses other approaches and their requirements.
 
To summarize, the main contributions of this paper are:
\begin{itemize}
    \item A novel, neuro-symbolic architecture (CPG) for sequence-to-sequence language tasks that is modular, recursive, and uses abstract grammar rules to ensure systematic behavior across classes of inputs.
    \item A simple, intuitive curricular training algorithm that allows CPG to incrementally learn the semantics for new types and rules.
    \item Perfect few-shot generalization to two popular compositional generalization benchmarks, SCAN and COGS using only 14 (out of ~14k) and 22 (out of ~24k) examples respectively. 
    \item Interpretability.
\end{itemize}



\begin{figure}
  \centering
  \includegraphics[width=\linewidth]{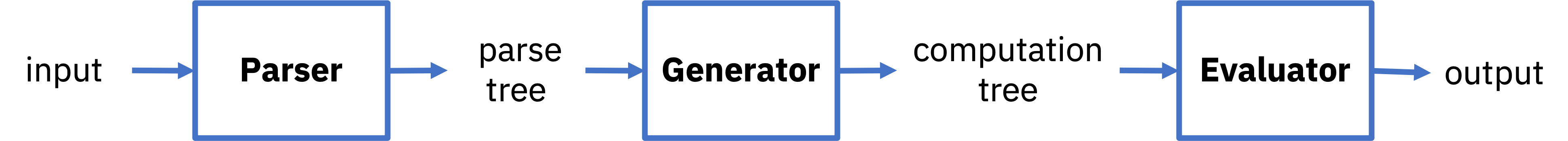}
  \caption{CPG architecture}
  \label{fig:cpg-arch}
\end{figure}

\section{Approach}
Our goal is to build a model capable of systematically composing pieces of the input that it already knows how to solve and handling arbitrarily long inputs. One class of functions which is well-suited to handle these requirements is the class of \textit{compositional} functions. These are often described informally as \textit{meaning} functions where \textit{the meaning of the whole input is a function of the meaning of parts (spans) of the input}. \footnote{Interestingly, although usually attributed to Frege, this idea appears in some form as early as the sixth century BCE \citep{Pagin2010-PAGCID}.} 

More formally, a \textit{compositional function} $\mu$ is a recursive function of the form:
\begin{equation*}
\mu(\alpha(u_1, u_2, \ldots, u_n)) = r(\alpha)(\mu(u_1), \mu(u_2), \ldots, \mu(u_n)).
\end{equation*}
 Here $u_1, u_2, \ldots, u_n$ are spans of the input which are the ``parts"; $\alpha$ is a composition function which combines these part spans into a larger ``whole" span $\alpha(u_1, u_2, \ldots, u_n)$; and $r(\alpha)$ is a semantic module which says how to combine the meanings of the parts $\mu(u_1), \mu(u_2)$, etc. For a formal presentation and additional details see \citep{Pagin2010-PAGCID}.

Compositional functions are a natural fit for compositional generalization for several reasons. First they are recursively defined, so support \textit{productive} generalization to arbitrarily long inputs. Second, they are \textit{systematic} in the sense that two expressions composed with the same rule $\alpha$ are consistently mapped to the same semantic module $r(\alpha)$, while expressions composed with different rules may be mapped to different modules. This means that a compositional function will consistently handle the entire class of expressions with the same compositional structure using the same composition of modules, but will allow different classes of expressions to be processed differently. By contrast, a monolithic model, such as a Transformer \citep{NIPS2017_3f5ee243}, processes all inputs with the same function regardless of how they are composed.

Our goal is to learn a compositional function by learning modules $r_{\theta}(\alpha)$ associated with context-free grammar rules $\alpha$. Figure \ref{fig:cpg-arch} shows the high-level architecture. We start by parsing the input using a supplied CFG and Lark\footnote{https://github.com/lark-parser/lark}.  After the input is parsed, the Generator takes the parse tree and produces a computation tree recursively. The bottom level is produced using a \textit{dictionary} to map the computation tree leaves to their corresponding expressions in the output language (for the experiments with SCAN, the dictionary is learned; for those with COGS it is supplied). The Generator then recursively maps the non-primitive grammar rules of the parse (e.g. $\delta$) to their associated modules (e.g. $r_{\theta}(\delta)$). We use two kinds of programs: \textit{copy} programs, which copy elements of their concatenated inputs to the output, and \textit{substitution} programs, which substitute objects into slots in the concatenated input expressions. Finally, the resultant computation tree is recursively evaluated to produce the output. The code for CPG (main algorithm, Generator, and Evaluator) is given in Algorithms \ref{fig:cpg-code}, \ref{fig:generate-code}, and \ref{fig:evaluate-code}, respectively. We illustrate the process with two examples.

\begin{algorithm}[t]
\caption{CPG -- Compositional Program Generator}
\label{fig:cpg-code}
    \begin{algorithmic}[1]
        \Function{$\texttt{CPG}_{\theta}$}{\texttt{g}, \texttt{x}}
           \State $\texttt{parser} \gets \texttt{create\_parser(g)}$
           \State $\texttt{parse} \gets \texttt{parser.parse(x)}$
           \State $\texttt{comp\_tree} \gets \texttt{generate}_{\theta}(\texttt{parse})$
           \State $\texttt{output} \gets \texttt{evaluate(comp\_tree)}$
        \EndFunction
    \end{algorithmic}
\end{algorithm}

\subsection{Example: SCAN}
\begin{figure}
  \centering
  \includegraphics[width=\linewidth]{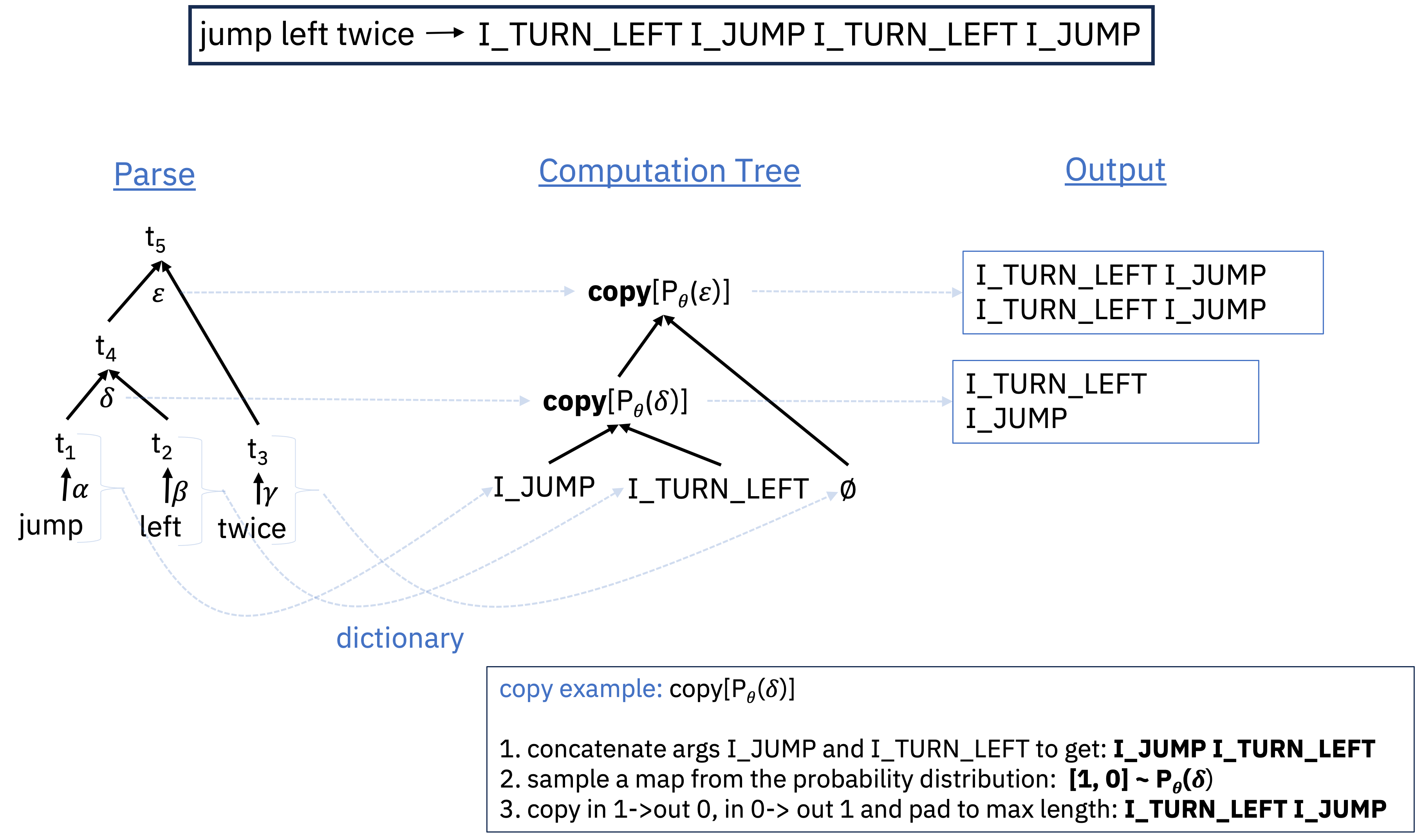}
  \caption{An example of compositional program generation on SCAN. Copy modules are parameterized by distributions dependent only on the abstract rules. An example of a copy operation is shown at the bottom. Also see Algorithm \ref{fig:copy}.}
  \label{fig:scan-example}
\end{figure}

In SCAN the task is to translate simplified English commands into a linear sequence of actions which could be executed by a robot on a grid. The primitive commands like \texttt{jump} and  \texttt{walk} map 1-1 to corresponding robot actions such as \texttt{I\_JUMP} or \texttt{I\_WALK}. These actions can be modified by directional terms such as \texttt{left} and \texttt{right} as well as terms indicating the number of times to repeat the command, such as \texttt{twice} or \texttt{thrice}. The full SCAN grammar is provided in Figure \ref{fig:grammars} (Left).

A parse of the input is shown in the left of Figure \ref{fig:scan-example}. We assume that each input token is mapped by a \textit{primitive rule} to its type in the grammar. For example, ``jump" is parsed by rule $\alpha$ to its primitive type $t_1$. \textit{Non-primitive} grammar rules apply to types not tokens directly, so for example $\delta$ parses the pair $(t_1, t_2)$ to the output type $t_4$.  The final reduction to the start symbol of the grammar is omitted.

The Generator (Algorithm \ref{fig:generate-code}) processes the parse in depth-first, postfix order (bottom-up, left-to-right), mapping each parse rule to a corresponding semantic module (shown to the right of the parse tree in Figure \ref{fig:scan-example}). Primitive grammar rules like $(\texttt{jump}, \alpha)$ and $(\texttt{left}, \beta)$ in the example are mapped by a \textit{dictionary} function (sometimes called a ``lexicon") to their corresponding meaning in the SCAN output language (or to the empty sequence). Non-primitive rules  like $\delta$ are mapped to probability distributions $P_{\theta}(\delta)$, with parameters $\theta$ to be learned. For SCAN, these distributions parameterize a symbolic \texttt{copy} program. We call the parameterized program a \textit{semantic module} (e.g. $\texttt{copy}[P_{\theta}(\delta)]$).

Finally, the computation tree is evaluated (Algorithm \ref{fig:evaluate-code}) in depth-first, post-fix order to produce the output show on the right in Figure \ref{fig:scan-example}. The \texttt{copy} program samples the distribution, applied to the input arguments, to get a map of input elements to output elements. An example of the operation of the the module $\texttt{copy}[P_{\theta}(\delta)]$, which has learned to switch its argument order, is shown at the bottom of the figure. The second copy module ($\textit{copy}[P_{\theta}(\epsilon)]$) appends a copy of its concatenated arguments to handle the ``twice" in the input and produce the final output.

In the implementation we assume that the parameters are private for each rule distribution, allowing them to specialize without interference and, for efficiency, we interleave generation of the semantic modules with evaluation to incrementally produce the output. (We've left off the subscripts on $\theta$ to avoid visual clutter). The example in figure \ref{fig:scan-example} shows the incremental output on the right. 

\subsection{Example: COGS}
\begin{figure}
  \centering
  \includegraphics[width=\linewidth]{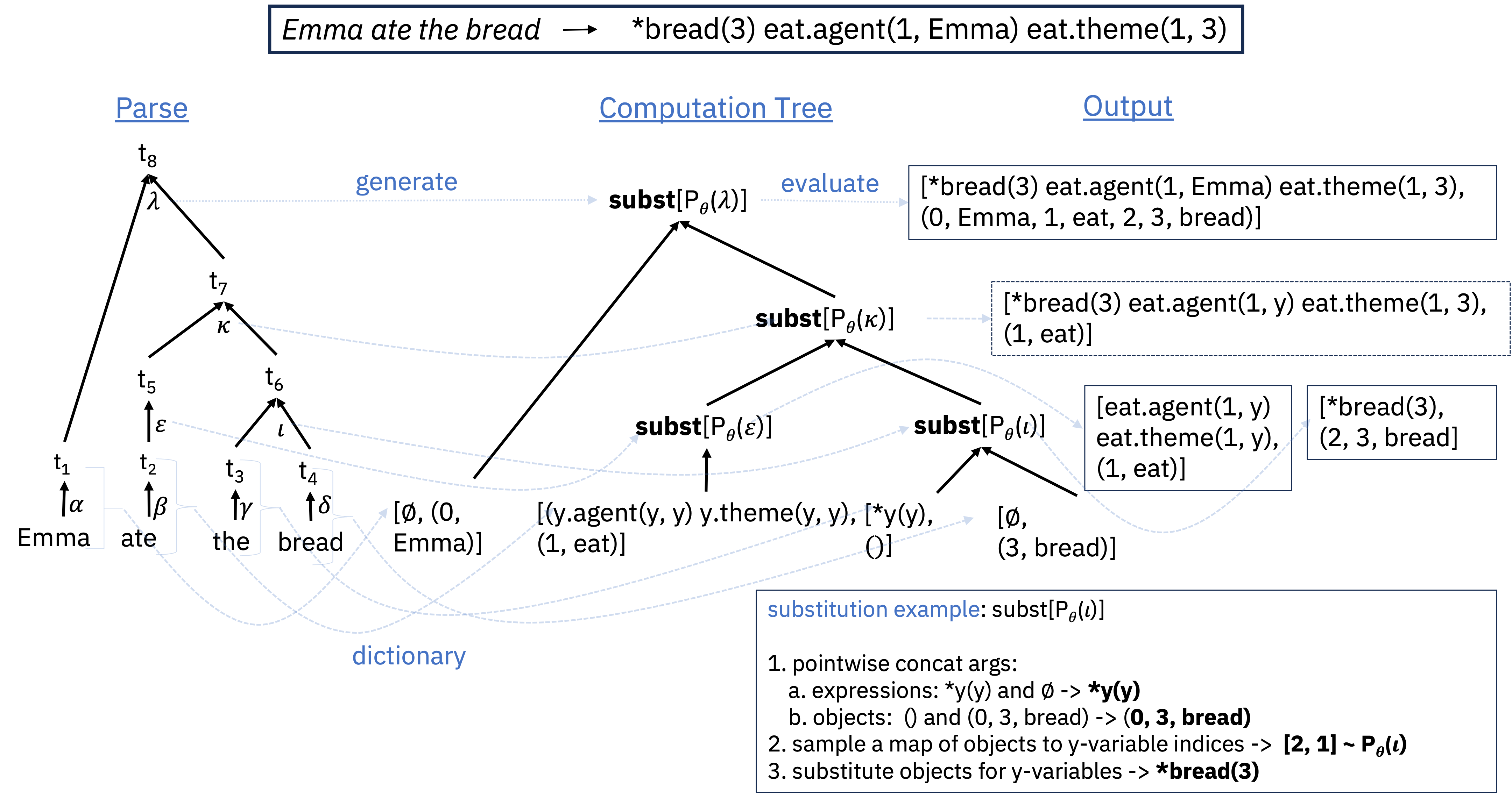}
  \caption{An example of compositional program generation on COGS. Substitution modules are parameterized by distributions dependent only on the abstract rules. An example of a substitution operation mapping objects to slots is shown at the bottom. Also see Algorithm \ref{fig:substitute}.}
  \label{fig:cogs_example}
\end{figure}

The COGS dataset requires translation of English sentences of arbitrary length to a logical form which captures their meaning. For example the sentence ``Emma ate the bread" would be represented logically as \texttt{*bread(3) eat.agent(1, Emma) eat.theme(1, 3)}. Here \texttt{*bread(3)} represents the fact that there is a bread object called \texttt{3} (the \texttt{*} means ``there exists"); the agent or subject who is eating is \texttt{Emma}; the event of Emma eating is denoted \texttt{1}; and the theme or object of the eating event is \texttt{3}, the bread. Other sentences can be translated to this form similarly. For a detailed description of COGS and formal logical language see \cite{kim2020cogs}. 

\begin{figure}
  \centering
  \includegraphics[width=\linewidth]{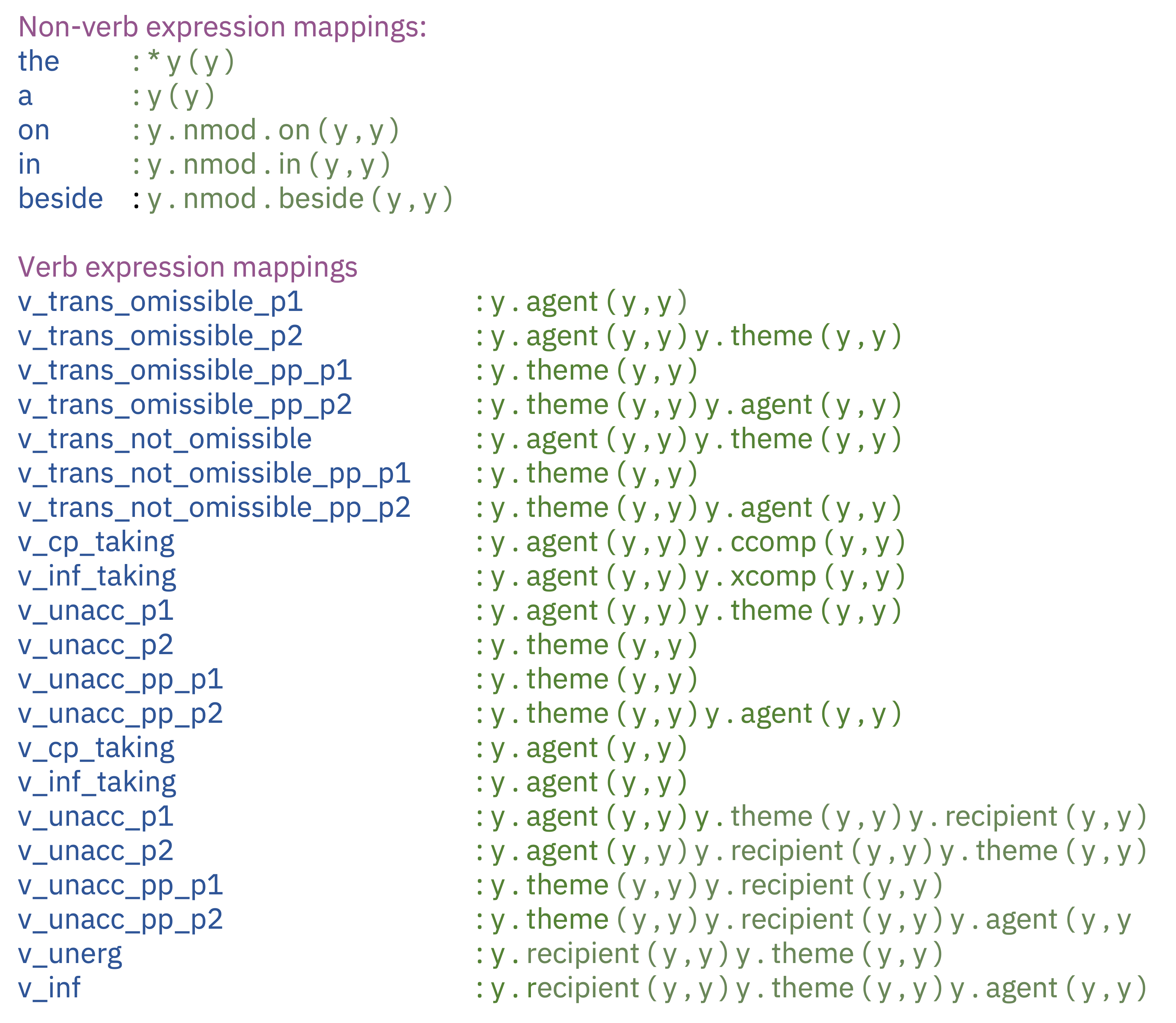}
  \caption{The COGS dictionary. Non-verb mappings depend on the token being mapped. Verb mappings depend on the primitive type of the verb. This is why the dictionary requires both the primitive type and token.}
  \label{fig:cogs-dictionary}
\end{figure}

\begin{figure*}[t]
  \centering
  \includegraphics[width=\linewidth]{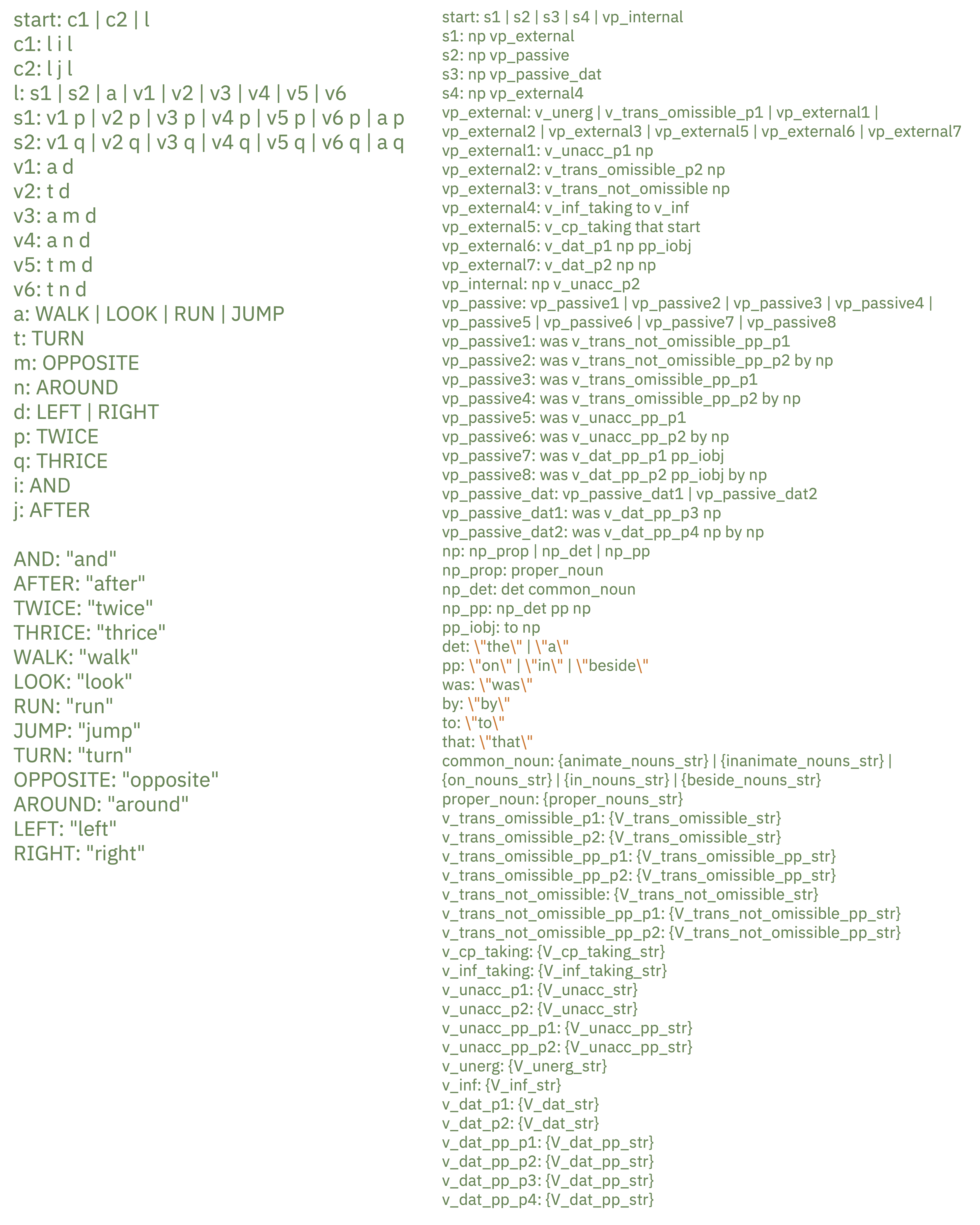}
  \caption{SCAN grammar (left); COGS grammar (right)}
  \label{fig:grammars}
\end{figure*}

Figure \ref{fig:cogs_example} shows how CPG processes the sentence ``Emma ate the bread". Again a Lark parser parses the input as shown in the parse tree on the left. The Generator uses the parse to recursively build the computation tree shown to the immediate right. The dictionary provides the translation of the individual input words (and their types) to an \textit{expression} which may contain variables or slots denoted $y$ (e.g. $y.agent(y, y)$)) and a list of \textit{objects} which can be variables (token indices) or constants (e.g. Emma).\footnote{The expressions for proper nouns are empty ($\phi$) since everything needed to interpret them is in the object list. For verbs like \texttt{ate} the expression includes predicates describing the subject (or agent), the object (or theme) and others, depending on type. The COGS dictionary is shown in Figure \ref{fig:cogs-dictionary}. The SCAN and COGS grammars are in Figure \ref{fig:grammars} for reference.} The Generator maps non-primitive rules to probability distributions which parameterize a substitution program shown in Algorithm \ref{fig:substitute}. An example of a substitution is shown at the bottom of Figure \ref{fig:cogs_example}.

\section{Model}
The model for CPG specified in Algorithms 1-5 requires learning the dictionary (in the case of SCAN) and the probability distributions associated with grammar rules, used in the \texttt{copy} and \texttt{substitute} programs.

\subsection{Dictionary distribution}
The dictionary used in SCAN computes a probability distribution over the output tokens: $P_{\theta}(\cdot|x)$ for $x$ a token in the input vocabulary. In the implementation we have $P_{\theta}(\cdot|x) = \sigma_G(FF_{\theta}(x))$ where $FF$ is a feed forward network with a single hidden layer of size 30 (by default) and $\sigma_G$ is a (hard) Gumbel softmax function \cite{jang2017categorical} \cite{maddison2017concrete}.

\subsection{Module distributions}
The probability distributions for the modules (one per grammar rule) each have private parameters. The distributions $P_{\theta}$ are computed as $\sigma_G(Linear_{\theta}(\mathbf(1))$, a (hard) Gumbel softmax $\sigma_G$ applied to a linear layer, which is parameterized by weights $\theta$ and applied to a vector of 1's.\footnote{Note that these distributions do not depend on the input directly, just rules. This is a feature that prevents over-fitting and forces generalization to whole classes of inputs}. The dimensions of the linear layer for the copy program match the maximum length of the CPG input/output vectors; and for substitution programs the maximum number of $y$ variables (slots) allowed in an expression.

To make the substitution and copy operations differentiable we do not use indexing to produce the output but instead compute a binary selector mask and multiply. Also, because the symbolic operations required to substitute and copy are complicated to formulate in a tensor style we use loops and do not batch the input, so one iteration corresponds one training example.

\begin{algorithm}[t]
\caption{Generate Computation Tree\\ 
MODULE may be \texttt{copy} or \texttt{subst}. \texttt{n.children} represents the children of node \texttt{n}; \texttt{n.rule} represents the rule that produced node \texttt{n}. The functions \texttt{add\_child} and \texttt{add\_children} append to this list of children.}
\label{fig:generate-code}
    \begin{algorithmic}[1]
        \Function{$\texttt{generate}_{\theta}$}{\texttt{n}}
            \If{\texttt{is\_primitive\_type(n)}}
	           \State $\texttt{leaf} \gets \texttt{n.child}$
	           \State $\texttt{return} \; \texttt{dictionary}_{\theta}\texttt{(n, leaf)}$
            \Else
                \State $\texttt{C} \gets [\texttt{generate}_\theta ( \texttt{c} )\; | \; \texttt{c} \in \texttt{n.children} ]$
                \State $\texttt{dist} \gets \texttt{P}_{\theta}(\texttt{n.rule})$
                \State $\texttt{module.function} \gets \texttt{MODULE}$
                \State $\texttt{module.add\_child(dist)}$
                \State $\texttt{module.add\_children(C)}$
                \State $\texttt{return} \; \texttt{module}$
            \EndIf
        \EndFunction
    \end{algorithmic}
\end{algorithm}
\begin{algorithm}[t]
\caption{Evaluate Computation Tree \\
\texttt{function.apply} applies \texttt{function} to the arguments.}
\label{fig:evaluate-code}
    \begin{algorithmic}[1]
        \Function{$\texttt{evaluate}$}{\texttt{n}}
            \If{\texttt{is\_leaf(n)}}
	           \State $\texttt{return} \; \texttt{n}$
            \Else
                \State $\texttt{P}_{\theta} \gets \texttt{n.children(0)}$
                \State $\texttt{x} \gets \texttt{concat}([\texttt{evaluate(n.children(i))} \; |$ \\
                $1 \leq \texttt{i} < |\texttt{n.children}|])$
                \State $\texttt{return}\; \texttt{n.function.apply}(\texttt{P}_{\theta}, \texttt{x})$
            \EndIf
        \EndFunction
    \end{algorithmic}
\end{algorithm}

\begin{algorithm}[t]
\caption{\texttt{Copy Program} \\
\texttt{pad} pads the output to the specified length.}
\label{fig:copy}
    \begin{algorithmic}[1]
        \Function{\texttt{Copy}}{\texttt{P, args}}
                \State $\texttt{x} \gets \texttt{concat(args)}$
                \State $\texttt{map} \gets \texttt{(y} \sim \texttt{P} \; | \; 1 \leq \texttt{i} \leq \texttt{|x|})$
                \State \texttt{return} \; \texttt{pad(x[map], |x|)}
        \EndFunction
    \end{algorithmic}
\end{algorithm}

\begin{algorithm}[t]
\caption{\texttt{Substitute Program} \\
\texttt{pointwise\_concat([[e, o], [e, o]) -> [[e, e], [o, o]]]}}
\label{fig:substitute}
    \begin{algorithmic}[1]
        \Function{\texttt{Substitute}}{\texttt{P, args}}
                \State $\texttt{expr, objs} \gets \texttt{pointwise\_concat(args)}$
                \State $\texttt{vars} \gets \texttt{expr.variables}$
                \State $\texttt{map} \gets (\texttt{k} \sim \texttt{P} \; | \; 1 \leq \texttt{i} \leq \texttt{|vars|})$
                \State $\texttt{result} \gets \texttt{expr.clone()}$
                \State $\texttt{result}[\texttt{vars}] \gets \texttt{objs[map]}$
                \State \texttt{return} \; \texttt{result, objs}
        \EndFunction
    \end{algorithmic}
\end{algorithm}

\section{Curricular Training}
We train the dictionary and module distributions with gradient descent optimization using the Adam optimizer with learning rate adjustment on plateaus. For each experiment, we run at least 5 runs with different random seeds and measure accuracy as the average proportion of exact matches to the target across these seeds.

\subsection{Objective}
Our objective is to find parameters $\theta$ which minimize the cross entropy of the $CPG_{\theta}$ function, supplied with a grammar $g$ and input $x$, with the true value $y$. Here $D$ represents the training data.

\begin{equation*}
\underset{\theta}{\texttt{argmin}}\underset{(x,y) \in D}{\mathbf{E}}[\texttt{cross\_entropy}(\texttt{CPG}_{\theta}(g, x), y)]
\end{equation*}

\subsection{Curriculum}
Training is curricular by stages made up of consecutive input sentence lengths. For SCAN the curriculum stages have are just single lengths: $1, 2, \ldots, 7$. For COGS the stages are 2-3, 4-6, 7-9, 10-12,  13-15, 16-18, and 19-21. At the beginning of each stage the stage accuracy is set to $0.0$ and the temperature for all Gumbel softmax distributions is set to $10.0$. 

As training proceeds the temperature is annealed proportional to the stage accuracy. When the stage accuracy reaches $1.0$, the maps sampled from each probability distribution $P_{\theta}(\alpha)$ are frozen for all rules $\alpha$ learned in that stage, and the model is evaluated on the validation set (if one is supplied) before moving to the next stage. The set of types needed to parse a sentence grows as sentence length increases so a curricular approach requires an incremental learning of the modules for any new rules. For COGS there are $60$ types in the grammar and all types appear somewhere in a parse of a sentence in each dataset. 

Training was performed on a MacBook Pro laptop without the use of GPUs.The code is open-source and publicly available here: https://github.com/IBM/cpg.

\section{Experiments}

\subsection{Construction of the Few-shot datasets}
\begin{figure}
  \includegraphics[width=\linewidth]{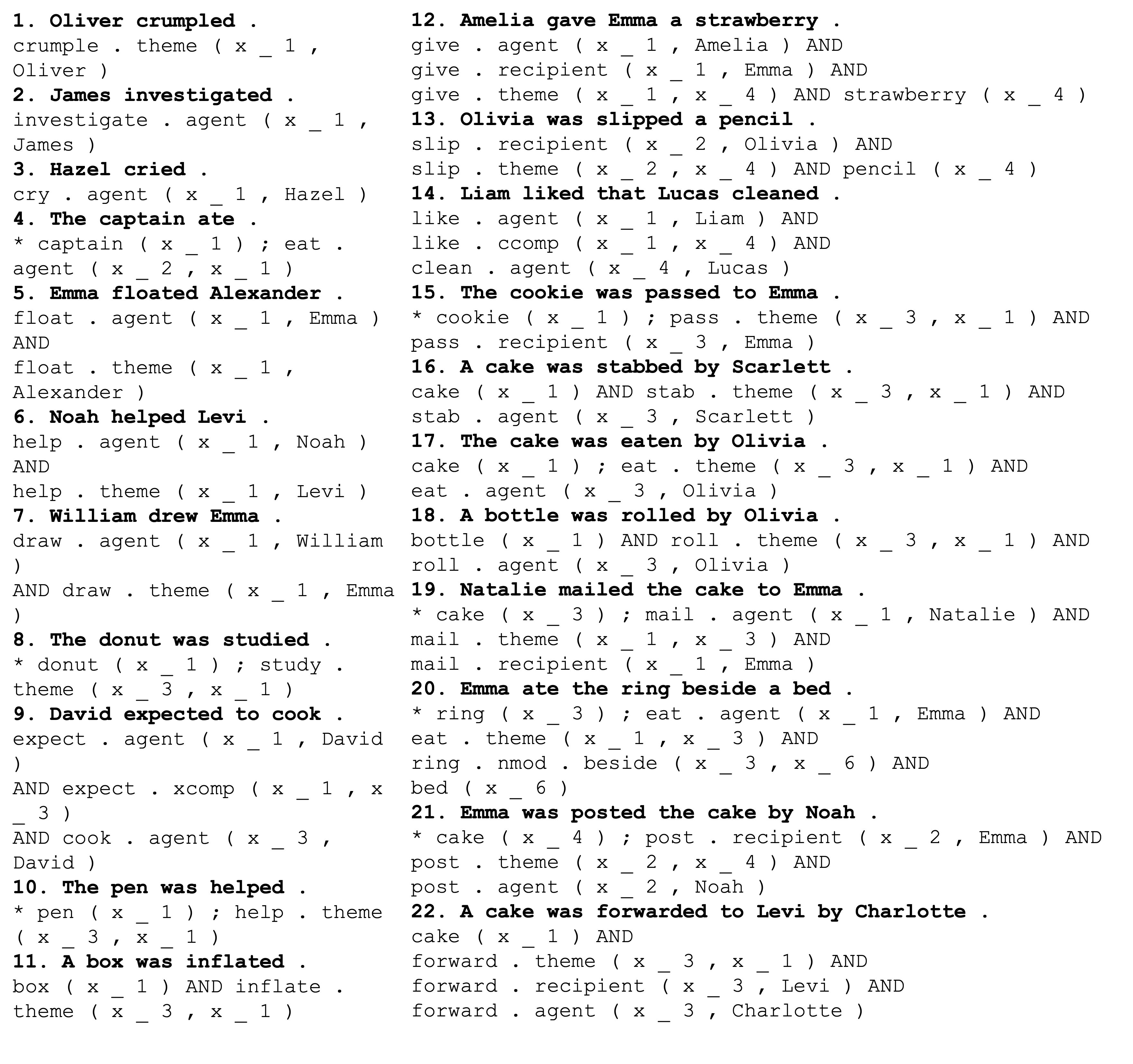}
  \caption{COGS few-shot instances}
  \label{fig:cogs-extreme-few-shot-instances}
\end{figure}

For both SCAN and COGS we produced few-shot versions of the training sets which are much reduced in size from the originals. For SCAN there are about $\sim 21k$ examples. The few-shot set used for both the length and add-jump splits has 14 examples; for COGS the original dataset has $\sim 24k$ examples, while  the few-shot training set contains just 22 examples, which are shown in Figure \ref{fig:cogs-extreme-few-shot-instances}.

To construct the datasets we sort the training set in ascending order by input length then loop over it, parsing each sentence, and keeping only those sentences which include new types (types not seen in previously parsed sentences). We have also experimented with just keeping sentences with \textit{parses} not previously seen in previous sentences, but there are more of these and the results are the same.

\subsection{Results and Comparisons}
\begin{figure}
  \includegraphics[width=\linewidth]{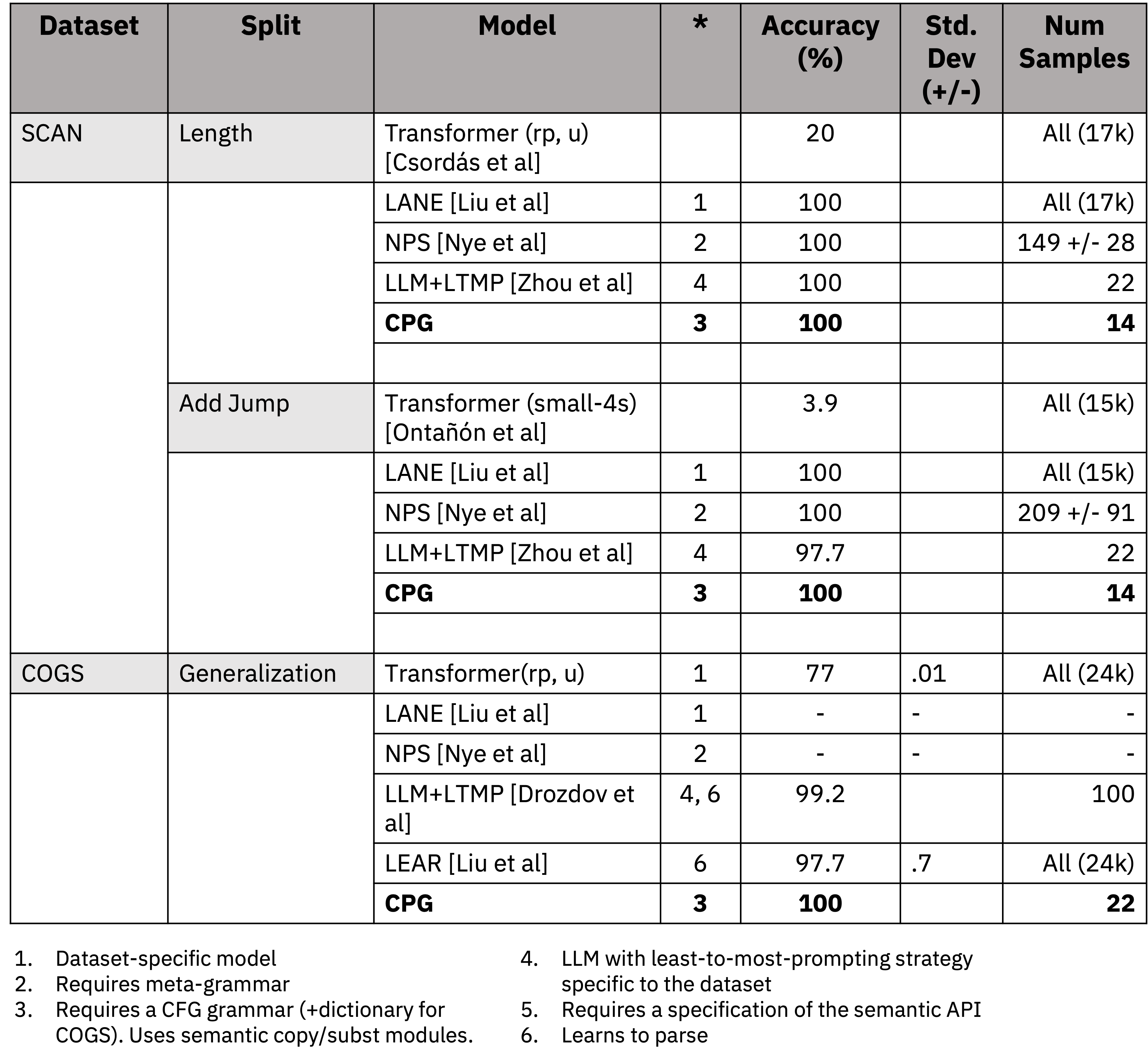}
  \caption{Comparison of Approaches}
  \label{fig:comparison}
\end{figure}

Figure \ref{fig:comparison} shows the generalization results for CPG and some of the relevant related approaches. It is difficult to directly compare these approaches because they make different assumptions and have different requirements. We have annotated the results in the (*) column to highlight important assumptions and more detail is provided on them below. A mark of (``-") indicates that no data is available for that split. 

For both SCAN (length and add-jump splits) and COGS (generalization split)  \textbf{CPG achieves perfect accuracy with several orders of magnitude less data} compared to standard approaches such as Transformers.

The best result on the SCAN length split from \citet{csordas-etal-2021-devil} use a universal Transformer with relative positional-encoding.\footnote{The authors conduct additional experiments with larger training lengths ($> 22$) and get perfect accuracy with a universal+relative encoding transformer setup when the length cutoff is extended to 25.} \cite{ontanon2022making} seems to have the best result for add-jump.

The LANE model of \citep{liu2020compositional} proposes a recursive and compositional approach that performs perfectly on all the splits of SCAN. However, their approach is non-trivial to adapt to other datasets like COGS. It's an end-to-end model which learns to parse and requires no human-supplied features.

The Neural Program Synthesis (NPS) approach of \cite{nye2020learning} is a neuro-symbolic meta-learning approach which learns to synthesize an interpretation grammar which fits the data, given a meta-grammar and a support set of a few hundred examples. It is more sample efficient than purely neural models but was not evaluated on COGS and it is unclear whether it would be able to scale. 

Approaches such as \cite{zhou2023leasttomost}, and \cite{drozdov2022compositional} use a pre-trained large language model with least-to-most-prompting (LTMP) to perform in-context ``learning". \cite{zhou2023leasttomost} provide the SCAN results which require a number examples of the same order of magnitude as CPG. These results were extended and enhanced to apply to the more challenging COGS dataset by \cite{drozdov2022compositional}, where accuracy is $99.2\%$. The COGS dataset has a small number of very difficult sub-problems (e.g. generalization to a length 55 sentence) so it is hard to know without a more detailed breakdown whether these problems lie in that $0.8\%$. Sample efficiency is roughly 5x worse than CPG on this task.

The LEAR approach of \citep{liu2021learning} performs well on COGS, among other datasets, learns to parse expressions, and does not require an input grammar or dictionary, but it does require semantic function definitions and does not learn in a modular way.

\begin{figure}
  \includegraphics[width=\linewidth]{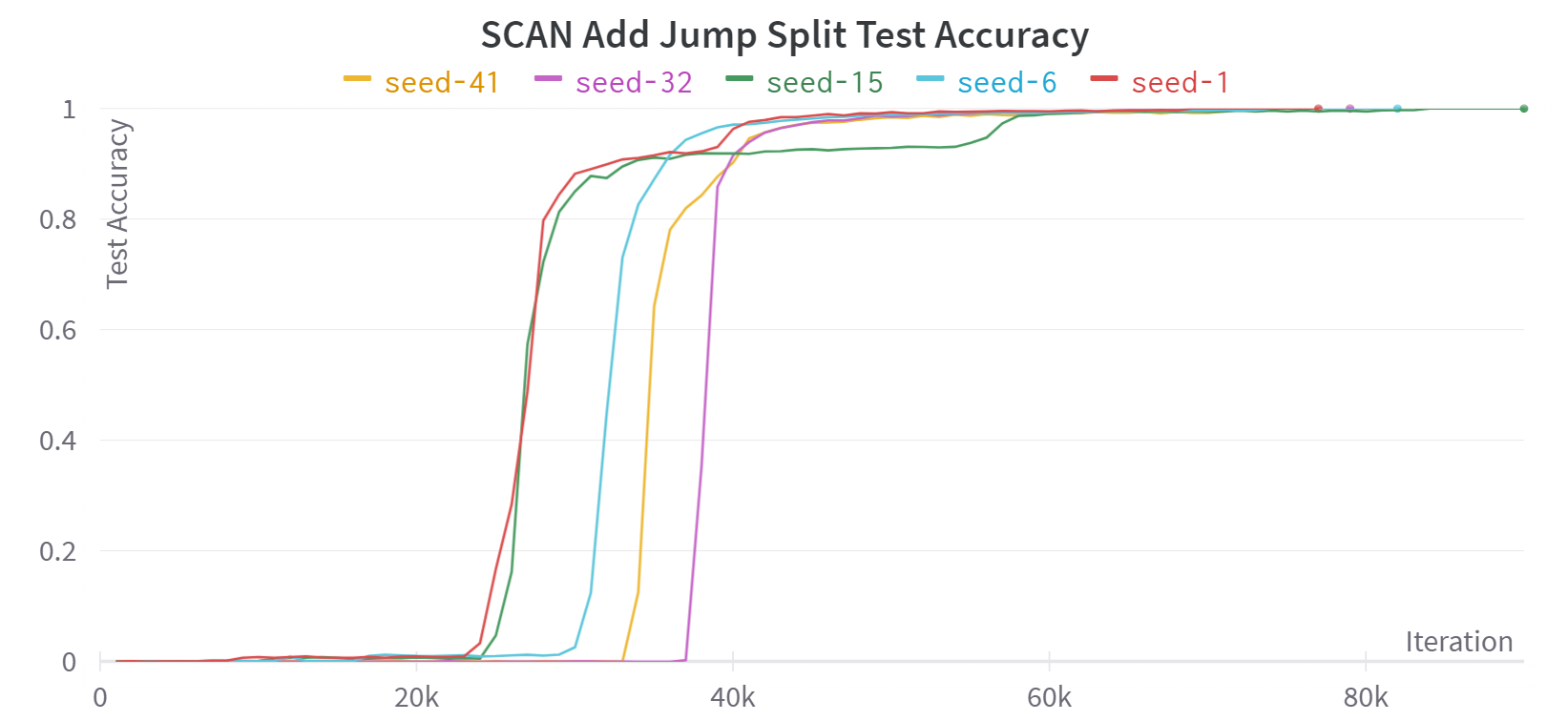}
  \includegraphics[width=\linewidth]{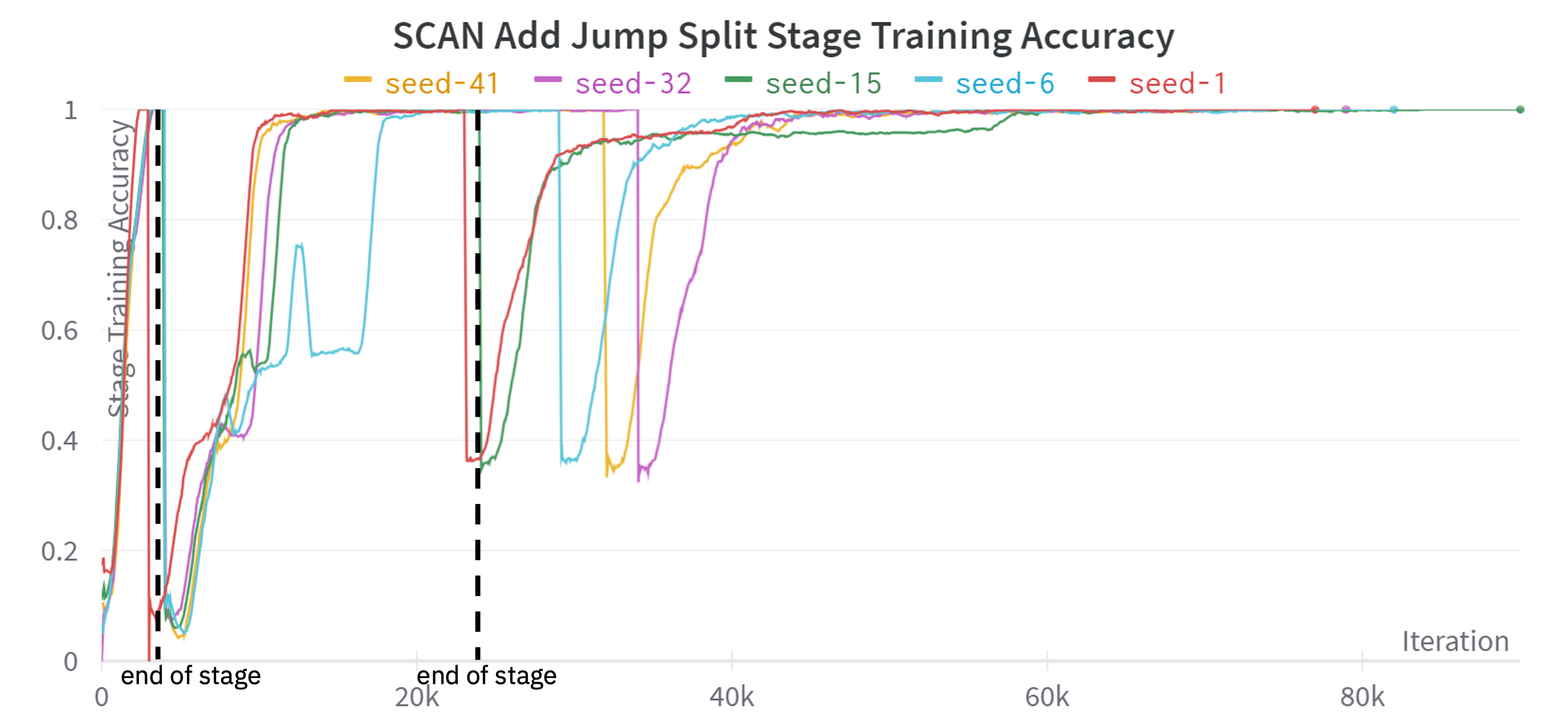}
  \caption{SCAN add-jump split. Top: test curve, Bottom: train curve. Results shown for 5 seeds. Vertical dash line indicates the end of the 1st and 2nd curriculum stages for seed 1.}
  \label{fig:scan-add-jump-training}
\end{figure}

To see how CPG is able to learn with less data we show the training curves for the SCAN add-jump split Figure \ref{fig:scan-add-jump-training}. The end of curricular stages 1 and 2 for seed 1 are indicated with a vertical dashed line. In each of these stage there is a rise in accuracy followed by a drop when the curriculum moves to the next stage and longer sentences with new types are introduced. Because the frozen sampled maps from module distributions learned in earlier stages are available to later stages, some longer sentences can be learned without any training at all (0-shot). This is why the accuracy does not go to 0 in the beginning of a new stage and in fact the drop decreases in later stages as more module distributions are learned.

\subsection{Incremental Re-Training}
Because the maps sampled for each module are frozen after a stage in which they are learned perfectly, we are able to retrain quickly without forgetting. This supports a use-case not typically available in purely neural models: efficient incremental re-training to address grammar issues or learn new concepts.

For example, when we constructed (or modified) context-free grammars for the COGS experiments we occasionally saw runs that got stuck in local minima. Because modules are linked to rules, we were able to quickly understand which small set of rules were involved. We found two issues in our hand-constructed grammars: (1) places in the grammar where sequences of types were repeated and could be replaced with a single type to avoid relearning the same concept multiple times independently and (2) places where two closely linked types exist which can be merged to reduce the search space (type merging). The need to split types is also a possible issue but we did not need to do it. One example occurred with the grammar snippet below where a COGS run performed sub-optimally at about 0.8 accuracy.
\begin{align*}
& \texttt{np\_det} \leftarrow \texttt{det} \; \texttt{common\_noun}\\
& \texttt{np\_p} \leftarrow \texttt{det} \; \texttt{common\_noun} \; \texttt{pp\_loc}\\
& \texttt{pp\_loc} \leftarrow \texttt{pp} \; \texttt{np}
\end{align*}
An examination of the types and rules involved showed that the map sampled from $P_{\theta}(\texttt{np\_pp})$, the distribution for the rule \texttt{np\_pp}, was incorrect. This is a sign that the module for this rule is finding it difficult to learn what is required. 

To remediate it, we refactored the grammar rule $\texttt{np\_pp} \leftarrow \texttt{det} \; \texttt{common\_noun} \; \texttt{pp\_loc}$ to replace the \texttt{det} and \texttt{common\_noun} elements with the \texttt{np\_det} type. This results in the grammar shown below.
\begin{align*}
& \texttt{np\_det} \leftarrow \texttt{det} \; \texttt{common\_noun}\\
& \texttt{np\_p} \leftarrow  \; \texttt{np\_det} \; \texttt{pp\_loc}\\
& \texttt{pp\_loc} \leftarrow \texttt{pp} \; \texttt{np}
\end{align*}
The intuition is that using the \texttt{np\_det} type is semantically equivalent to using \texttt{det} and \texttt{common\_noun} and if used in two contexts, gives the model training for the type in two contexts. Conversely, using \texttt{det} \texttt{common\_noun} in separate contexts requires the model to learn this concept twice independently. Because the model trains incrementally, re-training can start from the stage in which the error occurred. Accuracy improved after refactoring to about $0.92$ but there is another grammar optimization possible.

In the above grammar there is no reason to have \texttt{pp\_loc} and \texttt{np\_p} as separate types since \texttt{pp\_loc} has no special semantic meaning independent of \texttt{np\_p} and does not appear anywhere else in the grammar and makes it harder to learn. The solution is to remove the type \texttt{pp\_loc}:
\begin{align*}
& \texttt{np\_det} \leftarrow \texttt{det} \; \texttt{common\_noun}\\
& \texttt{np\_p} \leftarrow  \; \texttt{np\_det} \; \texttt{pp} \; \texttt{np}
\end{align*}

After applying both the type re-use and merge refactorings the performance becomes optimal. This procedure works primarily because of modularity: we can map between the performance of a module to a grammar rule. Although manual, it points to an intriguing meta-learning algorithm that would evaluate modules and make automatic adjustments to their corresponding grammar rules as a way to learn the grammar.

\section{Discussion}
What our results show is that with a correct grammar (and dictionary for COGS), CPG achieves perfect generalization on standard benchmarks in few-shot settings using two very simple and general modules for copying and substitution. When the grammar types are well-aligned with the semantic task the performance is very strong. When the grammar types are misaligned it can increase the search space for the distributions and impact performance, though the impact is localized.

Although CPG is directly applicable to problems for which the grammatical structure and dictionary is known, our aim is to scale it to a broader set of real-world problems. For this, factoring the problem into the sub-tasks of learning the grammar/parser, dictionary, and generator is a helpful decomposition which corresponds to cognitively coherent tasks: learn or be given a grammar, the meanings of words, and how to compose those meanings systematically to scale to novel combinations of words.

Human studies such as discussed in \citet{carey} provide evidence that people learn these different cognitive tasks with different strategies. For example, it is easier for people to learn to compose and reason with an existing hierarchy of concepts than it is for them to learn to refine their concepts when new information is provided.

Future work includes learning the grammar by using module performance to guide grammar changes and learning more challenging dictionaries, such as those required for COGS.


\section{Related Work}
The ability of neural networks to generalize compositionally has been the subject of a long-standing debate in cognitive science. \citep{fodor1988connectionism,Marcus2001TheAM}. Recently, there have been several benchmarks proposed that have renewed interest in the compositional generalization abilities of modern neural networks, demonstrating that such models still struggle with systematicity and productivity. 

\citet{lake2018generalization} introduce the SCAN dataset to test compositional generalization. They evaluate several sequence-to-sequence models, all of which generalize poorly. Follow-up work confirmed and strengthened these findings \citep{loula2018rearranging}. 

Similarly \citet{hupkes2020compositionality} propose the PCFG (probabilistic context-free grammar) task and evaluate LSTMs \citep{HochSchm97}, convolutional nets, and transformers, yielding mixed results, with all performing poorly on productivity tests. Later, \citet{csordas-etal-2021-devil} found that transformer performance could be improved with more careful tuning yielding increased performance on PCFG and COGS, though the performance was still poor on the CFQ (Compositional Freebase Questions) dataset of \citet{keysers2019measuring} and on the standard SCAN length split.

The COGS \citep{kim2020cogs} benchmark is designed to test model performance on semantic parsing in natural language \footnote{COGS is synthetic but uses everyday English words and has a relatively comprehensive recursive grammar with 60 distinct types.}. Again, they find that sequence-to-sequence encoder-decoder architectures fail to compositionally generalize.

Recently, new architectures have been proposed which help mitigate this problem. The NPS (Neural Program Synthesis), LANE (Learning Analytical Expressions), and LEAR (Learning Algebraic Recombination) approaches have been discussed in the Results and Comparison section.

\citet{herzig-berant-2021-span} propose an approach specifically for semantic parsing called SPAN\_BASED\_SP which uses a parser to recognize spans and generates programs for their semantics. Their approach is bottom-up, recursive, and predicts primitive type categories, but does not learn rule-specific modules and is not evaluated on COGS \footnote{They evaluate on a modified form of SCAN for semantic parsing, which differs from the original SCAN task.}.

Our approach rests on the functional definition of compositionality formalized and rigorously developed in \citet{Pagin2010-PAGCID}. 

The benefits of a modular approach to learning in natural language domains are well-documented in for example: \citet{andreas2017neural,rosenbaum2018routing,rosenbaum2019routing,cases2019recursive}. As far as we are aware, we are the first to apply a modular architecture to the problem of compositional generalization, and the use of modularity at the abstract rule level is novel here as well.

\section{Conclusion}
We have presented Compositional Program Generator (CPG), a novel neuro-symbolic architecture for sequential language problems, which has three key attributes that encourage systematic, productive and efficient learning: \textit{modularity}, in the form of programs parameterized by rule-specific, learned distributions, \textit{composition}, determined by a context-free grammar and consistent across whole classes of expressions, and \textit{abstraction} in the form of context-free grammar rules. Given an input language grammar and dictionary our implementation of CPG is able to achieve a new state-of-the-art result in solving the difficult COGS benchmark when reduced to just 22 exemplar sentences. Training is curricular, permits efficient re-training, and is low-variance when the grammar is well-factored and aligned with the semantic task.

\section{Acknowledgements}
We thank Najoung Kim and Tal Linzen for making the COGS grammar and dataset generator available, Joe Cappadona for his contributions to an earlier iteration of this project, and Ernest Davis for his support.

\clearpage

\bibliographystyle{named}
\bibliography{ijcai24}

\end{document}